\pdfoutput=1
\documentclass[11pt]{article}

\usepackage[final]{acl}

\usepackage{times}
\usepackage{latexsym}

\usepackage[T1]{fontenc}
\usepackage[utf8]{inputenc}

\usepackage{microtype}

\usepackage{inconsolata}

\usepackage{graphicx}

\usepackage{booktabs}

\usepackage{hyperref}
\usepackage{url}
\usepackage{verbatim}
\usepackage{multirow}
\usepackage{amsmath}
\usepackage{colortbl}
\usepackage{makecell}
\usepackage{placeins}
\usepackage{listings}

\title{Monte Carlo Temperature: a robust sampling strategy for LLM's uncertainty quantification methods}

\author{
Nicola Cecere\thanks{Equal contribution}, \quad 
Andrea Bacciu\footnotemark[1], \quad
Ignacio Fernández Tobías, \quad
Amin Mantrach \\
Amazon \\
\texttt{nicola.cecere@mail.polimi.it} \\
\texttt{\{andbac, tobiasi, mantrach\}@amazon.com} \\
}

\begin{document}

\maketitle

\begin{abstract}

Uncertainty quantification (UQ) in Large Language Models (LLMs) is essential for their safe and reliable deployment, particularly in critical applications where incorrect outputs can have serious consequences. Current UQ methods typically rely on querying the model multiple times using non-zero temperature sampling to generate diverse outputs for uncertainty estimation. However, the impact of selecting a given temperature parameter is understudied, and our analysis reveals that temperature plays a fundamental role in the quality of uncertainty estimates. The conventional approach of identifying optimal temperature values requires expensive hyperparameter optimization (HPO) that must be repeated for each new model-dataset combination. We propose Monte Carlo Temperature (MCT), a robust sampling strategy that eliminates the need for temperature calibration. Our analysis reveals that: 1) MCT provides more robust uncertainty estimates across a wide range of temperatures, 2) MCT improves the performance of UQ methods by replacing fixed-temperature strategies that do not rely on HPO, and 3) MCT achieves statistical parity with oracle temperatures, which represent the ideal outcome of a well-tuned but computationally expensive HPO process. These findings demonstrate that effective UQ can be achieved without the computational burden of temperature parameter calibration.

\end{abstract}

\section{Introduction}

Large Language Models (LLMs) have fundamentally transformed the way we interact with artificial intelligence, revolutionizing various domains, from content creation to complex problem-solving tasks \citep{DBLP:journals/corr/abs-2108-07258, 10.5555/3600270.3602070, orru_human-like_2023}. However, these powerful models can sometimes produce unreliable or incorrect outputs, raising concerns about their deployment in critical applications \citep{rohrbach-etal-2018-object, xiao-wang-2021-hallucination, bacciu-etal-2024-handling}. While significant research efforts have focused on improving LLMs' accuracy through techniques like Chain-of-Thought prompting \citep{10.5555/3600270.3602070} and Retrieval-Augmented Generation \citep{lewis2020retrieval}, parallel work has emerged on developing uncertainty quantification (UQ) methods to estimate model confidence as an indicator of potential errors \citep{kadavath_language_2022, kuhn_semantic_2023, lin_generating_2024}.

\begin{figure*}
    \centering
    \includegraphics[width=1\linewidth]{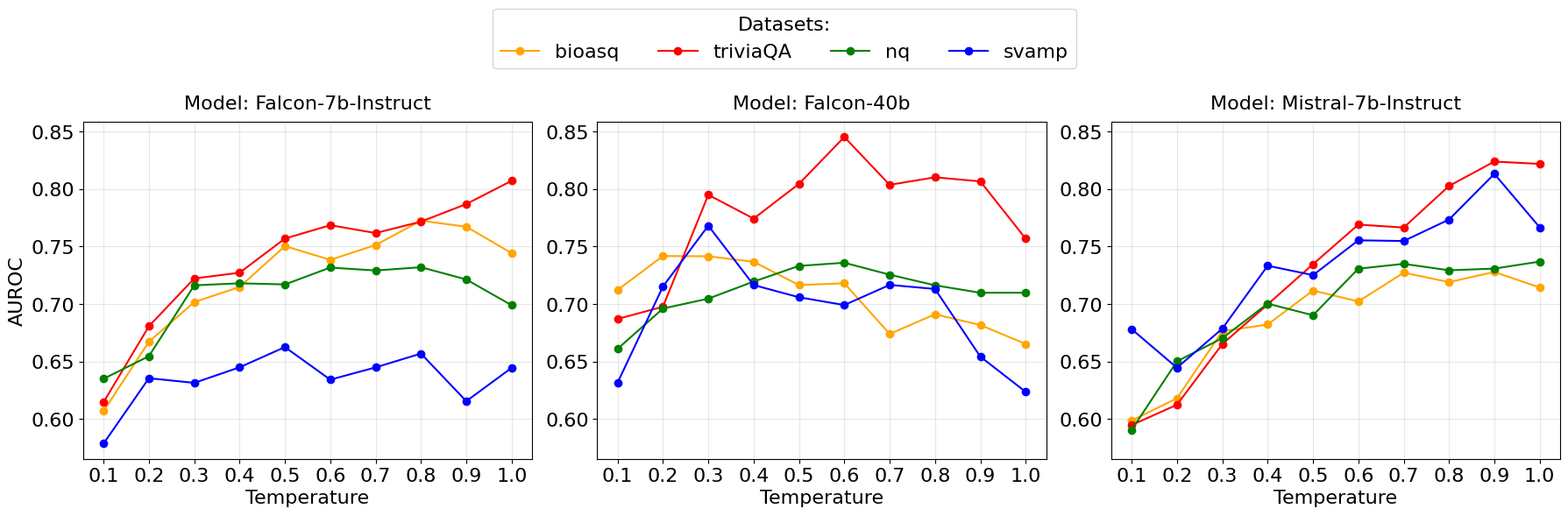}
    \caption{AUROC score distributions of the semantic entropy method across various model-dataset combinations and different fixed temperature values.}
    \label{fig:semantic_entropy_AUROC}
\end{figure*}

Existing UQ methods for LLMs can be used to predict the correctness of a LLM's output, either under white-blox or black-box assumptions. They fall into two broad categories: \textit{single-sample} and \textit{multi-sample} approaches.
\textit{Single-sample} methods analyze a single generation using metrics like perplexity or evaluating model's weight activations.
In contrast, \textit{multi-sample} methods, which we focus on in this work, rely on querying the model multiple times with the same input and non-zero fixed temperature sampling, to induce and measure diversity in the generations. To assess the effectiveness of UQ methods in distinguishing between correct and incorrect model outputs, they are typically evaluated as a classification procedure using the area under the receiver operator characteristic curve (AUROC) metric \citep{hanley1982meaning}.
However, the impact of selecting a specific fixed temperature parameter is understudied, and our analysis reveals that temperature plays a fundamental role in the effectiveness of different UQ methods across scenarios in which different LLMs are employed to solve different tasks.
Figure \ref{fig:semantic_entropy_AUROC} exemplifies this behavior over four question-answering datsets and three models using the semantic entropy method\footnote{Similar plots for other UQ methods can be found in the Appendix~\ref{sec:appendix}.} \citep{kuhn_semantic_2023}.
The figure highlights three critical observations: (1) for a given model and dataset, performance varies significantly with changes in temperature; (2) no single temperature consistently optimizes performance across datasets for a given model; and (3) no universal temperature yields optimal results across models for a given dataset.
For instance, the Falcon-40B model achieves peak performance on the TriviaQA dataset at a temperature of $0.6$, but requires a lower temperature of $0.3$ for the SVAMP dataset. Similarly, within the same TriviaQA dataset, optimal temperature values differ across different models: Falcon-40B performs best at $0.6$, while Falcon-7B-Instruct achieves superior results at $1.0$.
This lack of robustness in maintaining consistent performance across different scenarios poses significant challenges for practitioners attempting to implement UQ methods and highlight the need for more robust approaches to temperature selection.

To address the challenges of selecting a specific fixed temperature in UQ methods, we introduce \textit{Monte Carlo Temperature} (MCT), a sampling strategy that dynamically varies the temperature during multiple sentence generations, allowing UQ methods to generalize more effectively to different model-dataset combinations. This approach reduces sensitivity to specific temperature values and ensures more reliable uncertainty estimates.

We evaluate MCT against  an \textit{oracle} determined by selecting the temperature that yields the best results on the test set. By using an oracle as reference, we place ourselves in the most challenging evaluation scenario, as it represents an idealized outcome that hyperparameter optimization (HPO) may not achieve in practice.

Beyond this comparison, we assess MCT against two alternative model-dataset agnostic approaches, that do not require HPO: the \textit{Best On Average Temperature}, which selects a single fixed value performing well across multiple models and datasets, and the \textit{Fixed Random Temperature} approach that randomly chooses a single temperature.

Our results demonstrate that MCT consistently achieves statistical parity with the oracle, eliminating the need for expensive HPO. Additionally, MCT outperforms both the Best On Average Temperature and the Fixed Random Temperature strategies, further highlighting the benefits of structured temperature sampling.

The paper is structured as follows: in Section \ref{sec:multi-sample-uq-methods}, we present an overview of multi-sample UQ methods. In Section \ref{sec:robustness-mct-sampling}, we introduce the MCT approach and describe its implementation. Section \ref{sec:experimental-setup} details the experimental setup, including the LLMs, datasets, and evaluation metrics used. Section \ref{sec:results} presents the results of our experiments. Finally, in Section \ref{sec:conclusion}, we discuss the implications of our findings, acknowledge limitations, and outline potential future research directions.

\section{Multi-Sample UQ Methods}
\label{sec:multi-sample-uq-methods}

In this section, we present an overview of popular multi-sample UQ methods that we selected to evaluate the MCT sampling strategy. These methods represent a diverse set of approaches commonly employed for estimating uncertainty in LLMs.

\begin{itemize}

    \item \textbf{Naive Entropy (NE):} NE \citep{kuhn_semantic_2023} computes the uncertainty of model predictions by measuring the entropy of the generated output sequences based on their probabilities. For a given input \( x \), the probability of each output sequence \( y \) is computed using the chain rule of probability, which considers the joint probability of each token in the sequence. The entropy is then defined as:
    \begin{equation}
        H(x) = -\sum_{y \in S} \hat{p}(y|x) \log \hat{p}(y|x),
    \end{equation}

    where \( S \) represents the set of sampled sequences used for UQ.

    \item \textbf{Semantic Entropy (SE):} SE \citep{kuhn_semantic_2023} quantifies uncertainty by evaluating entropy across semantic clusters of the generated outputs. These clusters are formed based on semantic similarity, identified using an entailment model (as described in section \ref{subsec:evaluation-entailment-model}). For each cluster \( c \), the probability \( \hat{p}(c|x) \) is calculated by summing the probabilities of all sequences within the cluster, i.e., \( \hat{p}(c|x) = \sum_{y \in c} \hat{p}(y|x) \), where \( y \) represents a sequence assigned to cluster \( c \). Semantic entropy is then computed as:
    \begin{equation}   
    SE(x) = -\sum_{c \in C} \hat{p}(c|x) \log \hat{p}(c|x),
    \end{equation}
    where \( C \) represents the set of semantic clusters.

    \item \textbf{Discrete Semantic Entropy (DSE):} Unlike SE, DSE \citep{farquhar_detecting_2024} does not require model-provided probability scores. Instead, it approximates cluster probabilities using the relative frequency of samples within each cluster. This method is particularly effective in black-box settings where access to internal probability scores is restricted.

    \item \textbf{Number of Semantic Sets (NumSemSets):} NumSemSets \citep{lin_generating_2024} simplifies DSE by directly counting the number of unique semantic clusters identified by the entailment model, where a larger number of clusters indicates higher uncertainty in the model’s outputs.
    
    \item \textbf{P(True):} This technique \citep{kadavath_language_2022} is designed to capture the LLM’s uncertainty by structuring the task as a multiple-choice question. The LLM first generates a set of candidate answers based on a given prompt and then re-evaluates these responses by assigning probabilities. Specifically, the model is asked to determine whether a generated answer is correct by selecting between \textbf{\textit{True}} and \textbf{\textit{False}}, e.g., \textit{Is the possible answer: (A) True (B) False?}. The probability assigned to \textbf{(A)} is recorded as an uncertainty measure. A few-shot prompting strategy with examples from the training set is used to provide contextual guidance.

\end{itemize}

\section{Robustness and MCT Sampling for UQ}
\label{sec:robustness-mct-sampling}

In this section, we define the concept of robustness in the context of UQ methods and formalize the MCT sampling strategy.

\subsection{Robustness Definition in UQ Methods}
Robustness in the context of UQ refers to the stability and generalization of a UQ method's performance when applied across different settings. In our use case, robustness captures the range to which a UQ method remains effective in assessing uncertainty under changes in the following dimensions:
\begin{itemize}
    \item \textbf{Inference Parameters\footnote{In this work we focused on the study of the temperature parameter. Future work will focus on the other common generation parameters.}:} Variability in parameters such as temperature, top-k sampling, or nucleus sampling, which govern the stochastic nature of responses generated by LLMs.
    \item \textbf{Model Diversity:} Differences in architectures, training objectives, and scales of LLMs, requiring the UQ method to adapt without significant degradation in performance.
    \item \textbf{Dataset Variability:} Application to datasets with differing domains, topics, or complexity levels, ensuring the UQ method's efficacy across tasks.
\end{itemize}

\subsection{Monte Carlo Temperature}
MCT is a novel sampling strategy designed to improve robustness and avoid costly HPO by dynamically varying the temperature parameter across multiple queries for the same input. Traditional methods often rely on a fixed temperature value, $\tau$, selected through HPO. In contrast, MCT eliminates the need for HPO by introducing a probabilistic mechanism that samples temperature values from a predefined distribution.

MCT can be directly applied to any existing UQ multi-sample strategy. Instead of determining the ideal fixed temperature through extensive tuning, MCT dynamically samples temperatures, enabling the same UQ multi-sample method to perform robustly without additional optimization. This approach ensures that the method adapts seamlessly across varying model-dataset combinations.

The process of applying MCT to a query $x$ involves the following steps:
\begin{enumerate}
    \item Define a temperature distribution \(p(T)\) with support \([\tau_{\min}, \tau_{\max}]\), where \(\tau_{\min}\) and \(\tau_{\max}\) represent the minimum and maximum temperatures considered for sampling.
    \item Draw $k$ independent samples from the temperature distribution:
    \[
    \tau_i \sim p(T), \quad i \in {1, \dots, k}.
    \]
     \item Generate $k$ responses $y_i$ from a model $\mathcal{M}$, where each response is conditioned on the query $x$ and the corresponding sampled temperature $\tau_i$:
    \[
    y_i = \mathcal{M}(x; \tau_i), \quad i \in \{1, \dots, k\}.
    \]
    \item Apply the selected UQ multisample method based on the generated responses $\{y_1, y_2, \dots, y_k\}$.
\end{enumerate}

For this work, we used a discrete distribution with possible temperature values selected as equidistant points between the specified bounds $\tau_{\text{min}}$ and $\tau_{\text{max}}$. For a given number of generations $k$, the temperature values are drawn without replacement from the discrete set:
\begin{equation}
\{\tau_{\text{min}}, \tau_{\text{min}} + \Delta, \tau_{\text{min}} + 2\Delta, \dots, \tau_{\text{max}}\},
\label{eq:tau_sequence}
\end{equation}
where $\Delta = \frac{\tau_{\text{max}} - \tau_{\text{min}}}{k - 1}$.

\section{Experimental Setup}
\label{sec:experimental-setup}

This section outlines the experimental framework employed to evaluate the performance of MCT and related UQ methods. We detail the configurations used for answer generation, the LLMs and datasets selected for evaluation, and the specific entailment and evaluation models utilized in the study.

\subsection{Configuration for Generating Answers}

In this study, we applied UQ methods to the open question-answering task, focusing on sentence-length outputs. The temperature parameter for our experiments was sampled within the range $\tau_{\text{min}} = 0.1$ to $\tau_{\text{max}} = 1.0$. To ensure a balance between computational efficiency and statistical robustness, we generated $k=5$ outputs per question. Prior research has demonstrated that using $5$ generations provides results that closely approximate those obtained with $10$ generations \citep{farquhar_detecting_2024, lin_generating_2024}. 

Once the parameters $\tau_{\text{min}}$, $\tau_{\text{max}}$, and $k$ are defined, applying equation~\eqref{eq:tau_sequence} yields the exact interval that we employed for MCT sampling: $\{0.100, 0.325, 0.550, 0.775, 1.000\}$.

\subsection{LLMs and Datasets}
We evaluated the following LLMs: Falcon-7B-Instruct \citep{almazrouei2023falconseriesopenlanguage}, Mistral-7B-Instruct-v0.3 \citep{jiang2023mistral7b}, Falcon-40B \citep{almazrouei2023falconseriesopenlanguage}, and LLaMA-8B-Instruct-v3.1 \citep{grattafiori2024llama3herdmodels}. Note that due to the licensing of LLaMA models family, we accessed it via an API that provided text generations without likelihood scores.

Our experiments employed four open-question datasets covering various topics: TriviaQA \citep{joshi-etal-2017-triviaqa} and Natural Questions \citep{kwiatkowski-etal-2019-natural} for general knowledge, SVAMP \citep{DBLP:journals/corr/abs-2103-07191} for mathematics, and BIOASQ \citep{tsatsaronis_overview_2015} for biology.

We sampled 1,000 questions from each dataset, except for SVAMP, which contains fewer samples. In this case, all available questions were used. Notably, this represents a dataset size 2.5 times larger than that employed in the work of \citet{farquhar_detecting_2024}.

\subsection{Entailment and Evaluation Model} 
\label{subsec:evaluation-entailment-model}

This study employs semantic clustering to assess bidirectional entailment between pairs of answers, following the methodology outlined in \citet{farquhar_detecting_2024}. To implement it, we adopted an LLM-as-Judge approach, utilizing the Amazon Nova Micro \citep{Intelligence2024} model to perform clustering tasks.

For response correctness evaluation, we employed the LLM-as-Judge paradigm, a method proven to be more reliable than traditional substring-overlap metrics \citep{santilli2024on, zheng2023judgingllmasajudgemtbenchchatbot}. Claude Haiku 3.5 \citep{anthropic2024claude3} served as the evaluation model, configured to assess correctness based on the original question and reference answer in the dataset. To maintain consistency with \citet{farquhar_detecting_2024}, we ensured that correctness evaluation was conducted using an additional response generated with a fixed temperature of $0.1$. This setting minimizes randomness, producing more deterministic outputs that serve as a stable basis for evaluation.

Our evaluation framework mirrors the dual LLM-as-Judge structure employed in \citet{farquhar_detecting_2024}, where one model is dedicated to clustering and the other to correctness evaluation. However, while the original framework utilized GPT-3.5 for clustering and GPT-4 for evaluation \citep{brown2020language, openai2024gpt4technicalreport}, we relied on alternative LLMs.

To assess the effectiveness of UQ methods, we measured performance using AUROC, PR-AUC, and AURAC metrics \citep{hanley1982meaning, davis2006relationship, farquhar_detecting_2024}. Confidence intervals at the 95\% level were computed for all metrics via bootstrapping to ensure statistical relevance.

\section{Results}
\label{sec:results}

In this section, we present the results of the MCT sampling strategy, comparing its performance against three baselines: (1) the oracle temperature, selected to maximize test set performance, (2) the Best On Average Temperature across model-dataset combinations, and (3) the Fixed Random Temperature approach. First, we assess how closely MCT approximates the oracle temperature and achieves statistical parity. Then, we compare MCT to the two baselines that do not rely on HPO. Our results reveal that a previously optimal temperature does not necessarily generalize well across different model-dataset settings, as the Best On Average Temperature still underperforms relative to MCT. Meanwhile, the random baseline highlights the drawbacks of uninformed selection, showing that arbitrary temperature choices lead to unpredictable and often suboptimal results.

\subsection{Statistical Parity with Oracle Temperatures}

\begin{figure*}
    \centering
    \includegraphics[width=1\linewidth]{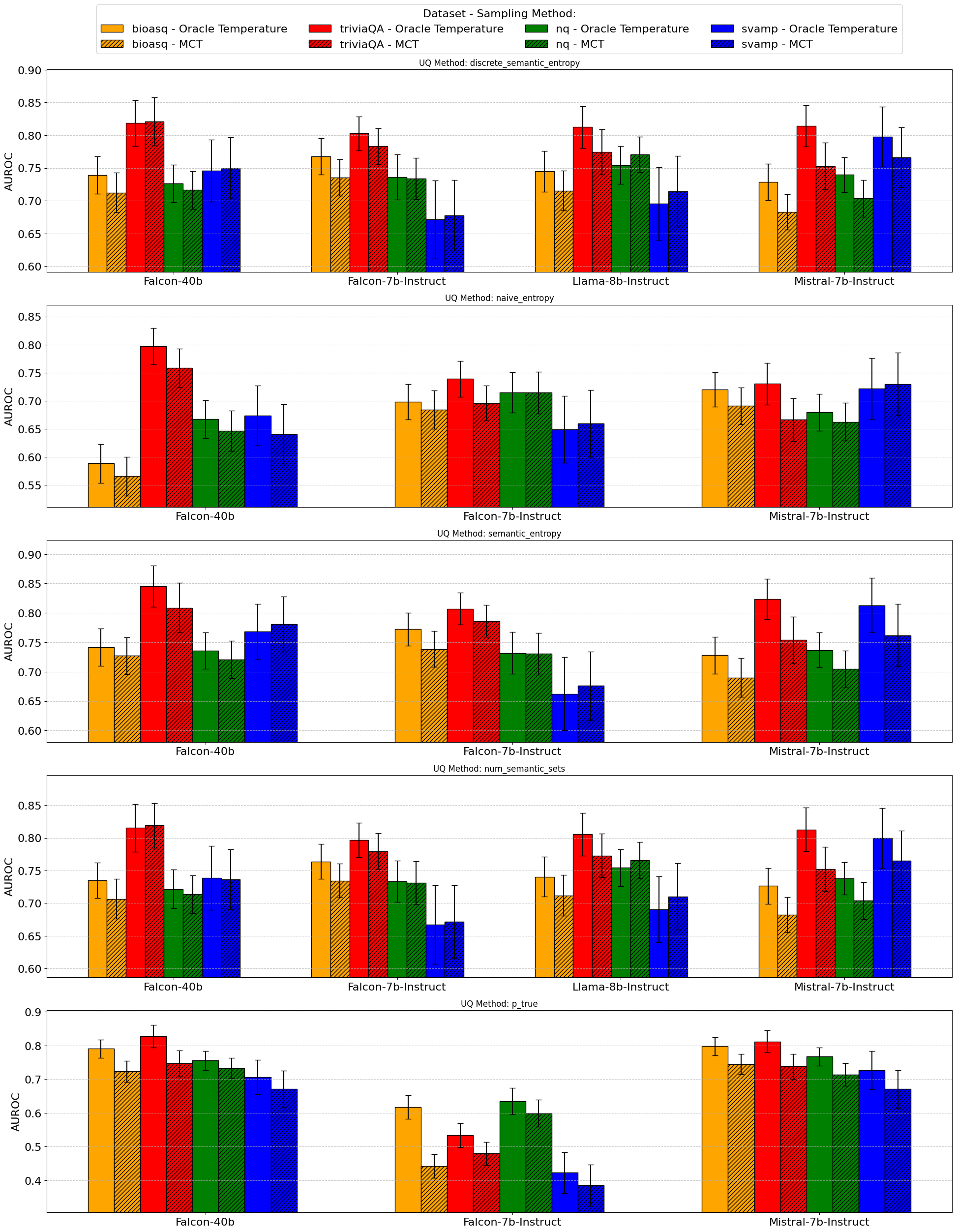}
    \caption{Comparison between oracle-fixed temperature performance and MCT sampling strategy performance across different UQ methods using the AUROC metric.}
    \label{fig:oracle_temperature_vs_mct_auroc}
\end{figure*}

Figure \ref{fig:oracle_temperature_vs_mct_auroc} demonstrates that MCT achieves statistical parity with optimal oracle-fixed temperatures across all UQ methods, models, and datasets, using statistical analysis at 95\% confidence level. This finding suggests that MCT can effectively replace any fixed temperature sampling approach while eliminating the need for temperature tuning. These results are further validated by additional performance metrics (PRAUC and AURAC), with detailed visualizations available in Appendix \ref{sec:appendix}.

\subsection{Comparison with the Best On Average and Fixed Random Temperature}

We evaluated MCT against a baseline approach that determined the best fixed temperature by averaging the scores obtained with each fixed temperature across all model-dataset combinations. To ensure an unbiased comparison, we applied leave-one-out cross-validation, systematically excluding each selected model along with all its associated datasets, as well as each selected dataset along with all its associated models, in the tested combination. The optimal temperature was then determined by averaging performance across the remaining combinations. This approach ensured that the test combination did not influence the temperature selection, effectively eliminating potential bias.

Additionally, we performed a comparison against a random baseline. To construct this baseline, we randomly sampled a fixed temperature 100 times from the same discrete range as MCT and computed the average performance across these simulations. This ensures a robust estimation of the expected performance when selecting a temperature at random, serving as an additional reference point for evaluating MCT’s effectiveness.

To assess performance, we quantified the relative difference, denoted as $\Delta$, which measures the deviation of each method (MCT, the best average fixed temperature, and the random baseline) from the oracle temperature’s performance. The results show that MCT consistently achieves a lower average $\Delta$ across all model-dataset configurations. Specifically, the average $\Delta$ for the best average fixed temperature method is $5.34\%$, while for the random baseline, it is higher at $5.85\%$. In contrast, MCT achieves an average $\Delta$ of $3.77\%$, demonstrating its superior adaptability and accuracy.

Moreover, this advantage translates into strong win-rate performance for MCT. It outperforms the Best Average Fixed Temperature method in 63.24\% of cases and achieves an even greater win rate of 72.03\% against the Random Baseline, further confirming its robustness.

Fine-grained results supporting these findings are provided in Table \ref{tab:combined_results_auroc} for the AUROC metric and in Appendix~\ref{sec:appendix} for the other metrics (PR-AUC, AURAC).

\newcommand{\g}{\cellcolor{gray!15}}

\begin{table*}[ht!]
\centering
\renewcommand{\arraystretch}{1.0}
\setlength{\tabcolsep}{5pt}
\resizebox{0.8\textwidth}{!}{%
\begin{tabular}{|c|l|c|c|c|c|c|c|c|}
\hline
\multicolumn{9}{|c|}{\textbf{Discrete Semantic Entropy}} \\
\hline
\textbf{Model} & \textbf{Dataset} & \textbf{Oracle} & \textbf{MCT} & \textbf{Best Avg.} & \textbf{Random} & \textbf{MCT $\Delta$ (\%)} & \textbf{Best Avg. $\Delta$ (\%)}  & \textbf{Random $\Delta$ (\%)} \\
\hline
\multirow{4}{*}{\centering Falcon-7b-Instruct} & triviaQA & 0.8028 & 0.7832 & 0.7880 & 0.7453 & 2.44 & \textbf{1.84} & 7.16 \\
& \g bioasq & \g 0.7681 & \g 0.7353 & \g 0.7639 & \g 0.7178 & \g 4.27 & \g \textbf{0.54} & \g 6.55 \\
& svamp & 0.6713 & 0.6778 & 0.6307 & 0.6267 & \textbf{-0.97} & 6.05 & 6.64 \\
& \g nq & \g 0.7361 & \g 0.7339 & \g 0.7282 & \g 0.7036 & \g \textbf{0.29} & \g 1.07 & \g 4.41 \\
\hline
\multirow{4}{*}{\centering Mistral-7b-Instruct} & triviaQA & 0.8143 & 0.7528 & 0.7326 & 0.7217 & \textbf{7.55} & 10.03 & 11.38 \\
& \g bioasq & \g 0.7286 & \g 0.6829 & \g 0.7226 & \g 0.6880 & \g 6.27 & \g \textbf{0.82} & \g 5.57 \\
& svamp & 0.7981 & 0.7662 & 0.7604 & 0.7192 & \textbf{3.99} & 4.73 & 9.88 \\
& \g nq & \g 0.7397 & \g 0.7036 & \g 0.6937 & \g 0.6923 & \g \textbf{4.88} & \g 6.22 & \g 6.40 \\
\hline
\multirow{4}{*}{\centering Falcon-40b} & nq & 0.7262 & 0.7164 & 0.6966 & 0.7025 & \textbf{1.34} & 4.07 & 3.25 \\
& \g triviaQA & \g 0.8185 & \g 0.8208 & \g 0.7882 & \g 0.7733 & \g \textbf{-0.28} & \g 3.71 & \g 5.52 \\
& svamp & 0.7462 & 0.7498 & 0.6255 & 0.6718 & \textbf{-0.49} & 16.17 & 9.96 \\
& \g bioasq & \g 0.7394 & \g 0.7125 & \g 0.6617 & \g 0.6966 & \g \textbf{3.64} & \g 10.51 & \g 5.79 \\
\hline
\multirow{4}{*}{\centering Llama-8b-Instruct} & triviaQA & 0.8125 & 0.7746 & 0.8023 & 0.7872 & 4.66 & \textbf{1.25} & 3.11 \\
& \g nq & \g 0.7544 & \g 0.7708 & \g 0.7517 & \g 0.7407 & \g \textbf{-2.17} & \g 0.35 & \g 1.81 \\
& bioasq & 0.7450 & 0.7155 & 0.7142 & 0.7197 & 3.96 & 4.13 & \textbf{3.40} \\
& \g svamp & \g 0.6957 & \g 0.7144 & \g 0.6957 & \g 0.6637 & \g \textbf{-2.69} & \g 0.00 & \g 4.59 \\
\hline
\multicolumn{9}{|c|}{\textbf{Naive Entropy}} \\
\hline
\multirow{4}{*}{\centering Falcon-7b-Instruct} & triviaQA & 0.7391 & 0.6959 & 0.6960 & 0.7021 & 5.85 & 5.84 & \textbf{5.00} \\
& \g bioasq & \g 0.6983 & \g 0.6842 & \g 0.6865 & \g 0.6768 & \g 2.03 & \g \textbf{1.70} & \g 3.08 \\
& svamp & 0.6489 & 0.6595 & 0.6157 & 0.6258 & \textbf{-1.64} & 5.12 & 3.56 \\
& \g nq & \g 0.7147 & \g 0.7145 & \g 0.7075 & \g 0.6895 & \g \textbf{0.04} & \g 1.02 & \g 3.54 \\
\hline
\multirow{4}{*}{\centering Mistral-7b-Instruct} & triviaQA & 0.7303 & 0.6663 & 0.7016 & 0.6556 & 8.77 & \textbf{3.93} & 10.23 \\
& \g bioasq & \g 0.7201 & \g 0.6907 & \g 0.7057 & \g 0.6852 & \g 4.08 & \g \textbf{1.99} & \g 4.84 \\
& svamp & 0.7215 & 0.7298 & 0.7050 & 0.6933 & \textbf{-1.16} & 2.28 & 3.90 \\
& \g nq & \g 0.6794 & \g 0.6626 & \g 0.6606 & \g 0.6576 & \g \textbf{2.47} & \g 2.77 & \g 3.20 \\
\hline
\multirow{4}{*}{\centering Falcon-40b} & nq & 0.6670 & 0.6464 & 0.6414 & 0.6499 & 3.10 & 3.84 & \textbf{2.57} \\
& \g triviaQA & \g 0.7973 & \g 0.7587 & \g 0.7692 & \g 0.7665 & \g 4.85 & \g \textbf{3.53} & \g 3.86 \\
& svamp & 0.6733 & 0.6406 & 0.5901 & 0.6338 & \textbf{4.85} & 12.36 & 5.86 \\
& \g bioasq & \g 0.5882 & \g 0.5656 & \g 0.5415 & \g 0.5614 & \g \textbf{3.86} & \g 7.94 & \g 4.57 \\
\hline
\multicolumn{9}{|c|}{\textbf{Semantic Entropy}} \\
\hline
\multirow{4}{*}{\centering Falcon-7b-Instruct} & triviaQA & 0.8072 & 0.7861 & 0.7716 & 0.7348 & \textbf{2.61} & 4.40 & 8.97 \\
& \g bioasq & \g 0.7725 & \g 0.7386 & \g 0.7725 & \g 0.7208 & \g 4.39 & \g \textbf{0.00} & \g 6.69 \\
& svamp & 0.6626 & 0.6763 & 0.6343 & 0.6356 & \textbf{-2.06} & 4.27 & 4.09 \\
& \g nq & \g 0.7320 & \g 0.7305 & \g 0.7320 & \g 0.7045 & \g 0.21 & \g \textbf{0.00} & \g 3.76 \\
\hline
\multirow{4}{*}{\centering Mistral-7b-Instruct} & triviaQA & 0.8239 & 0.7538 & 0.7347 & 0.7276 & \textbf{8.50} & 10.83 & 11.69 \\
& \g bioasq & \g 0.7279 & \g 0.6900 & \g 0.7023 & \g 0.6907 & \g 5.21 & \g \textbf{3.52} & \g 5.11 \\
& svamp & 0.8132 & 0.7620 & 0.7554 & 0.7396 & \textbf{6.29} & 7.10 & 9.05 \\
& \g nq & \g 0.7369 & \g 0.7046 & \g 0.7294 & \g 0.6935 & \g 4.38 & \g \textbf{1.03} & \g 5.90 \\
\hline
\multirow{4}{*}{\centering Falcon-40b} & nq & 0.7359 & 0.7209 & 0.7098 & 0.7133 & \textbf{2.05} & 3.55 & 3.08 \\
& \g triviaQA & \g 0.8452 & \g 0.8090 & \g 0.8102 & \g 0.7742 & \g 4.29 & \g \textbf{4.15} & \g 8.40 \\
& svamp & 0.7682 & 0.7808 & 0.6543 & 0.6888 & \textbf{-1.64} & 14.82 & 10.34 \\
& \g bioasq & \g 0.7418 & \g 0.7271 & \g 0.6818 & \g 0.7085 & \g \textbf{1.98} & \g 8.09 & \g 4.48 \\
\hline
\multicolumn{9}{|c|}{\textbf{Number of Semantic Sets}} \\
\hline
\multirow{4}{*}{\centering Falcon-7b-Instruct} & triviaQA & 0.7966 & 0.7795 & 0.7871 & 0.7305 & 2.14 & \textbf{1.20} & 8.30 \\
& \g bioasq & \g 0.7638 & \g 0.7346 & \g 0.7624 & \g 0.7283 & \g 3.82 & \g \textbf{0.18} & \g 4.65 \\
& svamp & 0.6669 & 0.6720 & 0.6215 & 0.6299 & \textbf{-0.76} & 6.81 & 5.56 \\
& \g nq & \g 0.7336 & \g 0.7313 & \g 0.7265 & \g 0.7062 & \g \textbf{0.32} & \g 0.98 & \g 3.74 \\
\hline
\multirow{4}{*}{\centering Mistral-7b-Instruct} & triviaQA & 0.8127 & 0.7526 & 0.7085 & 0.7416 & \textbf{7.40} & 12.82 & 8.74 \\
& \g bioasq & \g 0.7265 & \g 0.6826 & \g 0.7189 & \g 0.6760 & \g 6.05 & \g \textbf{1.05} & \g 6.95 \\
& svamp & 0.7994 & 0.7654 & 0.7606 & 0.7327 & \textbf{4.26} & 4.85 & 8.34 \\
& \g nq & \g 0.7380 & \g 0.7038 & \g 0.6952 & \g 0.6892 & \g \textbf{4.64} & \g 5.80 & \g 6.62 \\
\hline
\multirow{4}{*}{\centering Falcon-40b} & nq & 0.7215 & 0.7138 & 0.6920 & 0.6996 & \textbf{1.07} & 4.09 & 3.03 \\
& \g triviaQA & \g 0.8153 & \g 0.8191 & \g 0.7809 & \g 0.7724 & \g \textbf{-0.46} & \g 4.21 & \g 5.26 \\
& svamp & 0.7388 & 0.7365 & 0.6160 & 0.6589 & \textbf{0.31} & 16.61 & 10.81 \\
& \g bioasq & \g 0.7349 & \g 0.7067 & \g 0.6567 & \g 0.6950 & \g \textbf{3.83} & \g 10.65 & \g 5.43 \\
\hline
\multirow{4}{*}{\centering Llama-8b-Instruct} & triviaQA & 0.8056 & 0.7728 & 0.8024 & 0.7877 & 4.08 & \textbf{0.39} & 2.22 \\
& \g nq & \g 0.7542 & \g 0.7658 & \g 0.7506 & \g 0.7393 & \g \textbf{-1.54} & \g 0.48 & \g 1.98 \\
& bioasq & 0.7405 & 0.7120 & 0.7085 & 0.7188 & 3.85 & 4.32 & \textbf{2.93} \\
& \g svamp & \g 0.6907 & \g 0.7104 & \g 0.6907 & \g 0.6602 & \g \textbf{-2.86} & \g 0.00 & \g 4.41 \\
\hline
\multicolumn{9}{|c|}{\textbf{P(True)}} \\
\hline
\multirow{4}{*}{\centering Falcon-7b-Instruct} & triviaQA & 0.5335 & 0.4796 & 0.4924 & 0.4858 & 10.11 & \textbf{7.72} & 8.95 \\
& \g bioasq & \g 0.6170 & \g 0.4421 & \g 0.5442 & \g 0.5398 & \g 28.33 & \g \textbf{11.80} & \g 12.51 \\
& svamp & 0.4228 & 0.3852 & 0.3802 & 0.3941 & 8.89 & 10.07 & \textbf{6.78} \\
& \g nq & \g 0.6352 & \g 0.5990 & \g 0.6232 & \g 0.6024 & \g 5.71 & \g \textbf{1.90} & \g 5.17 \\
\hline
\multirow{4}{*}{\centering Mistral-7b-Instruct} & triviaQA & 0.8122 & 0.7383 & 0.7417 & 0.7680 & 9.09 & 8.68 & \textbf{5.44} \\
& \g bioasq & \g 0.7983 & \g 0.7445 & \g 0.7532 & \g 0.7564 & \g 6.73 & \g 5.65 & \g \textbf{5.25} \\
& svamp & 0.7273 & 0.6709 & 0.6540 & 0.6848 & 7.76 & 10.09 & \textbf{5.85} \\
& \g nq & \g 0.7672 & \g 0.7137 & \g 0.7342 & \g 0.7393 & \g 6.97 & \g 4.30 & \g \textbf{3.63} \\
\hline
\multirow{4}{*}{\centering Falcon-40b} & nq & 0.7556 & 0.7330 & 0.6575 & 0.6899 & \textbf{2.99} & 12.98 & 8.69 \\
& \g triviaQA & \g 0.8282 & \g 0.7469 & \g 0.8005 & \g 0.7915 & \g 9.82 & \g \textbf{3.35} & \g 4.43 \\
& svamp & 0.7070 & 0.6713 & 0.5797 & 0.6583 & \textbf{5.05} & 18.00 & 6.88 \\
& \g bioasq & \g 0.7906 & \g 0.7234 & \g 0.7208 & \g 0.7573 & \g 8.50 & \g 8.83 & \g \textbf{4.22} \\
\hline
\end{tabular}%
}
\caption{Performance comparison of UQ methods using AUROC score. Bold values show best performance per scenario, with $\Delta$ indicating difference from oracle baseline (lower $\Delta$ is better). Note: MCT $\Delta$ may be negative when performance exceeds the oracle baseline.}
\label{tab:combined_results_auroc}
\end{table*}

\section{Conclusion}
\label{sec:conclusion}

In this work, we introduced MCT, a general and robust sampling method for UQ in LLMs. Our approach eliminates the need for expensive HPO of temperature parameters, providing consistent performance across a wide range of models, datasets, and UQ methods. The experimental results demonstrate that MCT achieves statistical parity with oracle-fixed temperatures obtained through computationally intensive optimization. Additionally, it outperforms the best average fixed-temperature and random baselines by reducing performance variability and enhancing robustness across diverse configurations.

MCT's flexibility makes it applicable to any UQ method requiring multiple generations, and its dynamic temperature sampling effectively addresses challenges associated with fixed temperature configurations. This adaptability highlights MCT as a practical solution for deploying UQ methods in real-world scenarios where computational resources are limited.

\section{Limitations}

While this study demonstrates promising results, several limitations must be acknowledged. Although we validated MCT across a diverse set of UQ techniques and LLMs, further exploration is required to assess its effectiveness on larger-scale models and alternative architectures. Additionally, this work primarily focused on temperature as the inference parameter; future studies should examine the impact of other sampling techniques and inference configurations, such as top-P and top-k sampling, to expand MCT's applicability.

\section{Acknowledge}
We want to thank Marcello Federico for his valuable support and feedback on this paper.

\FloatBarrier

\bibliography{custom}

\FloatBarrier

\appendix
\section{Appendix}
\label{sec:appendix}

\begin{figure*}
    \centering
    \includegraphics[width=1\linewidth]{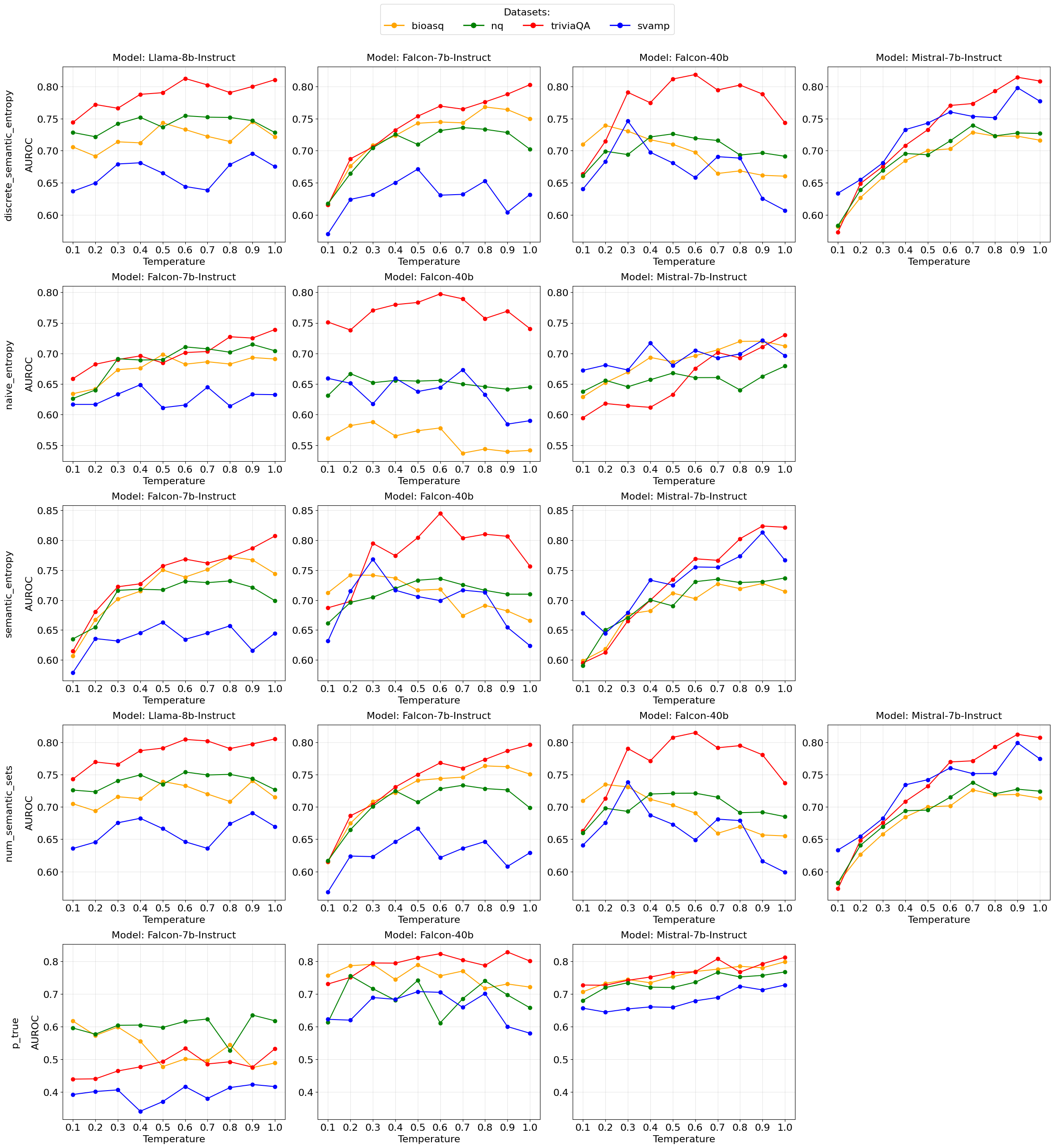}
    \caption{AUROC score distributions of tested UQ methods across various model-dataset combinations at different fixed temperature values.}
    \label{fig:temperature_distribution_AUROC}
\end{figure*}

\begin{figure*}
    \centering
    \includegraphics[width=1\linewidth]{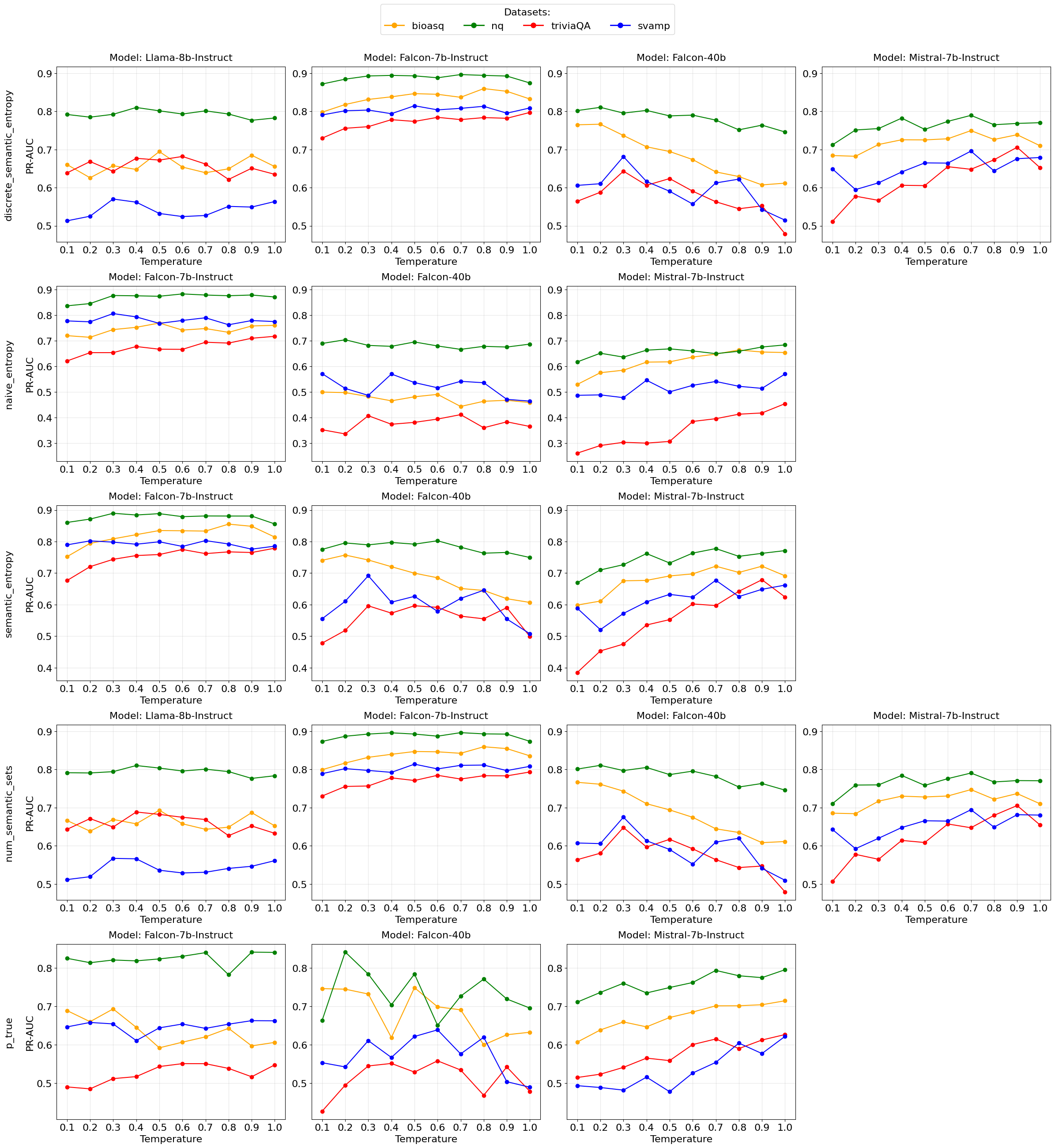}
    \caption{PR-AUC score distributions of tested UQ methods across various model-dataset combinations at different fixed temperature values.}
    \label{fig:temperature_distribution_PR-AUC}
\end{figure*}

\begin{figure*}
    \centering
    \includegraphics[width=1\linewidth]{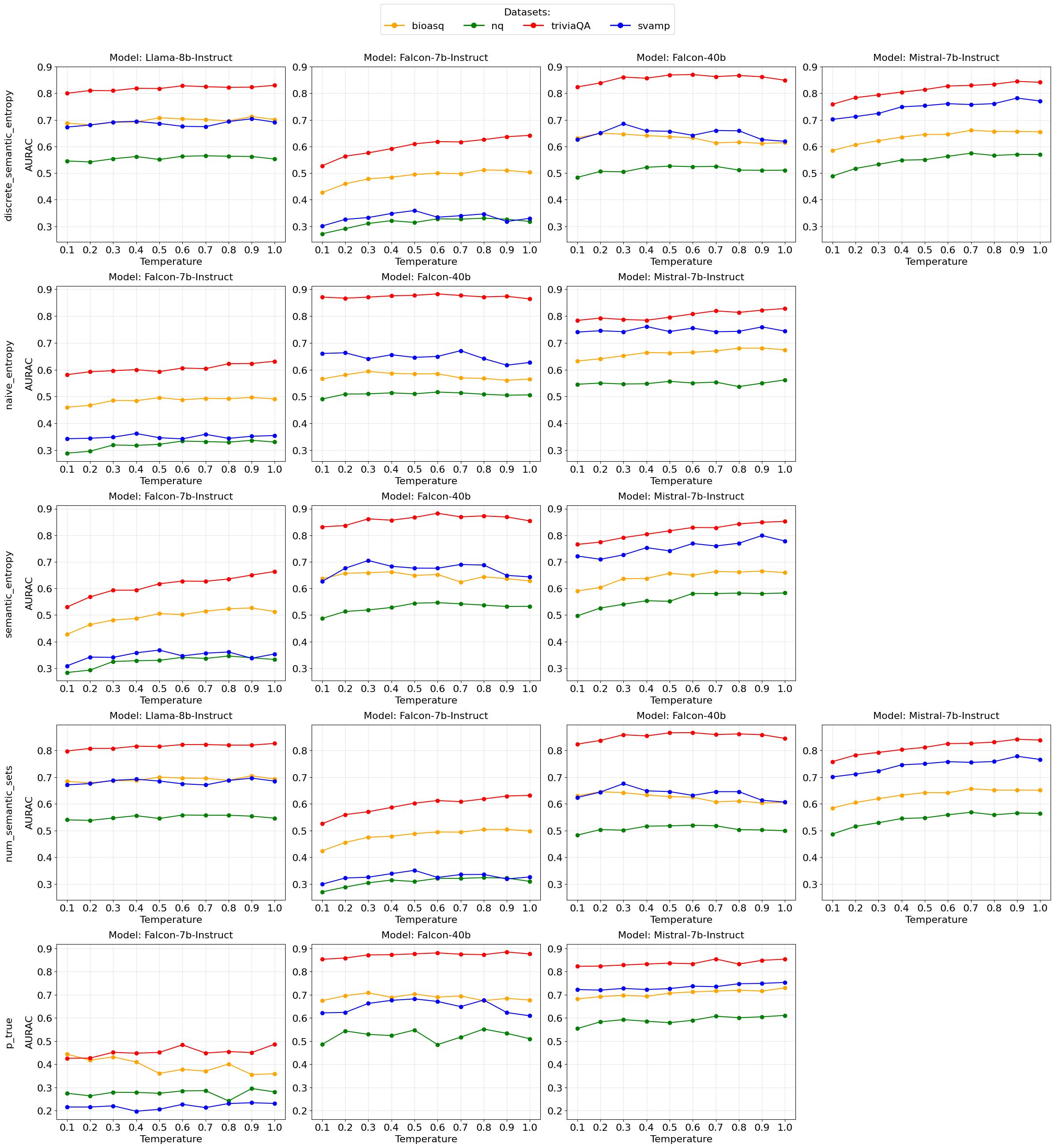}
    \caption{AURAC score distributions of tested UQ methods across various model-dataset combinations at different fixed temperature values.}
    \label{fig:temperature_distribution_AURAC}
\end{figure*}

\begin{figure*}
    \centering
    \includegraphics[width=1\linewidth]{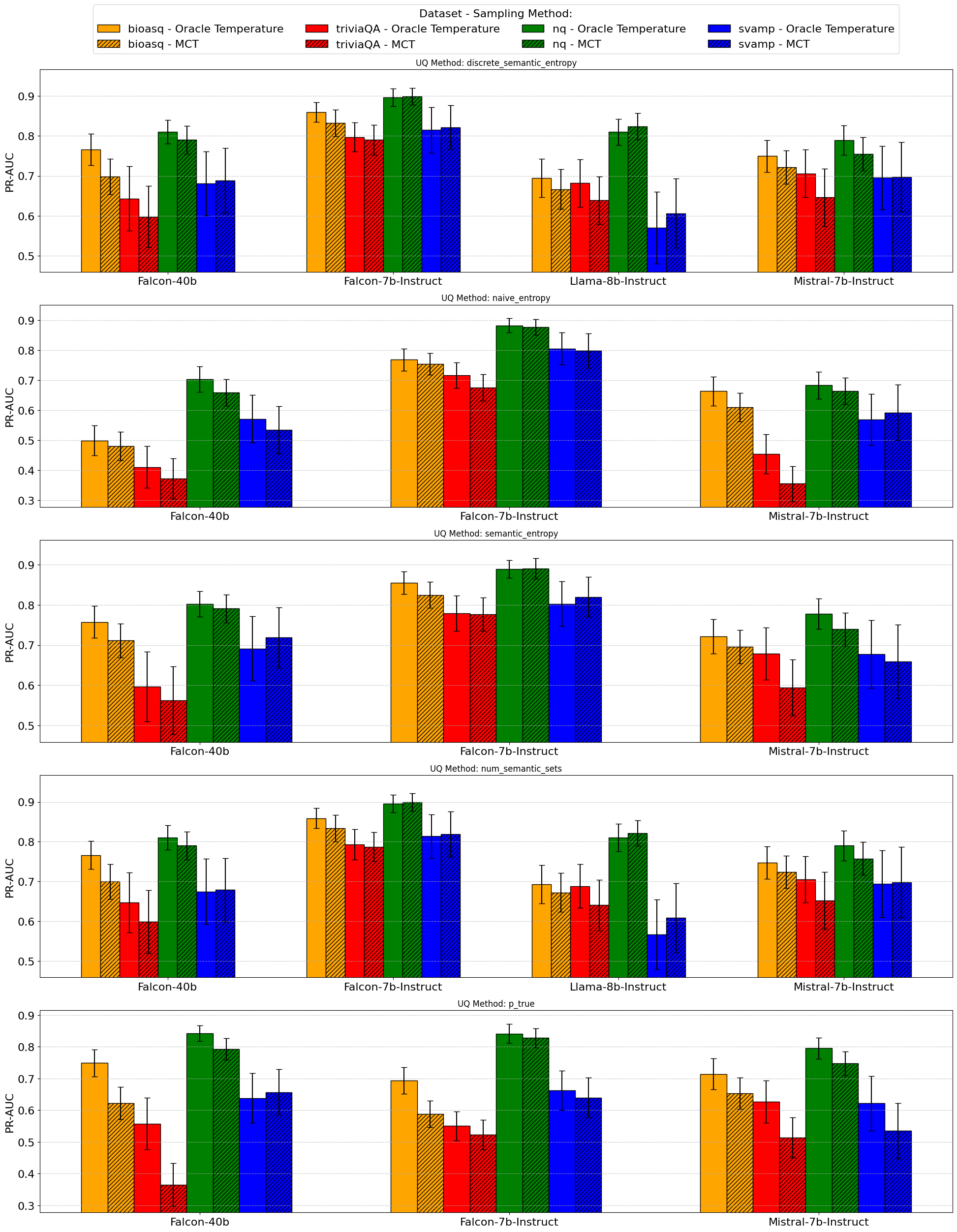}
    \caption{Comparison between oracle-fixed temperature performance and MCT sampling strategy performance across different UQ methods using the PR-AUC metric.}
    \label{fig:oracle_temperature_vs_mct_prauc}
\end{figure*}

\begin{figure*}
    \centering
    \includegraphics[width=1\linewidth]{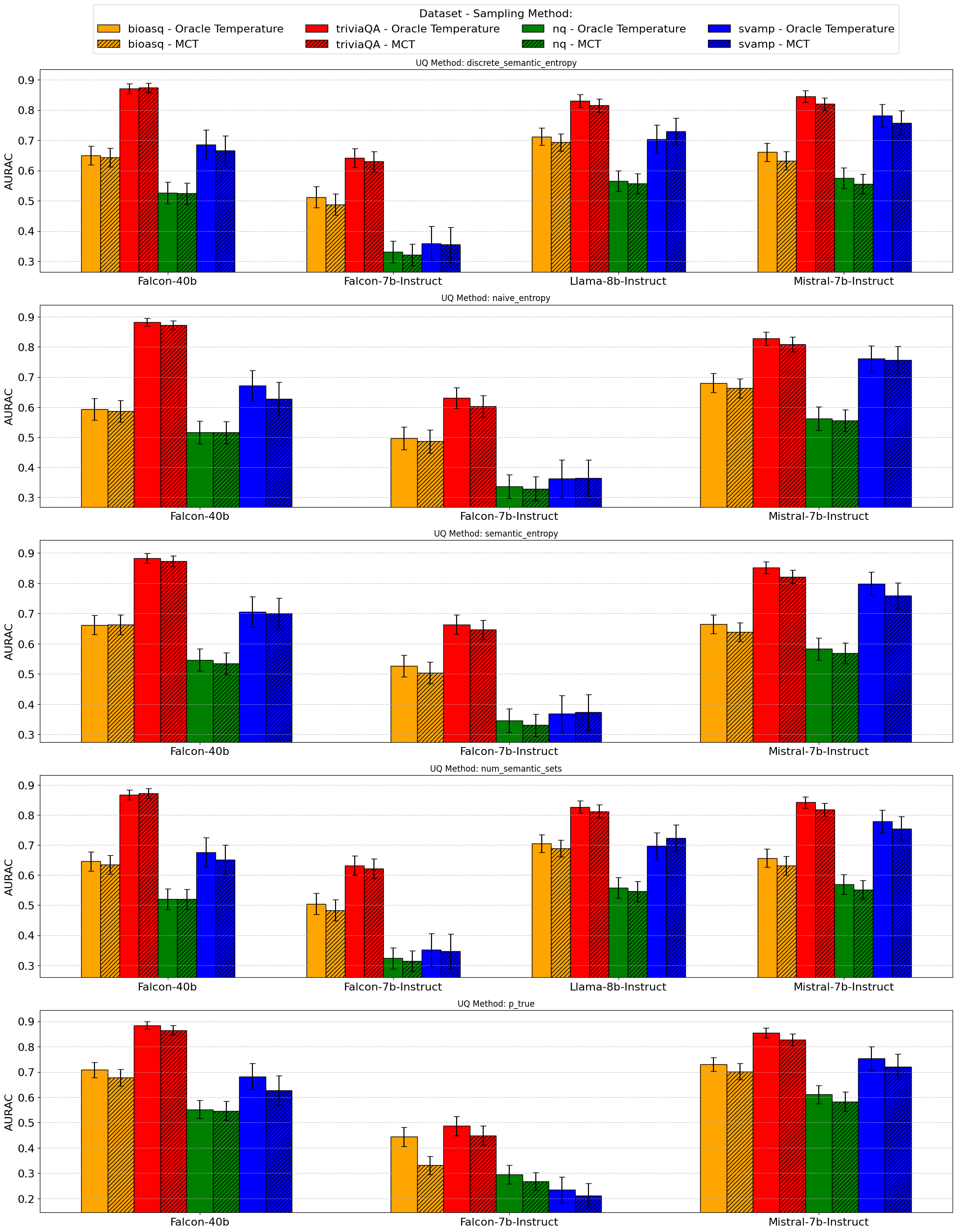}
    \caption{Comparison between oracle-fixed temperature performance and MCT sampling strategy performance across different UQ methods using the AURAC metric.}
    \label{fig:oracle_temperature_vs_mct_aurac}
\end{figure*}

This appendix provides additional quantitative results supporting the findings presented in the main text. The following figures and tables illustrate the performance of MCT compared to the oracle temperature and non-HPO fixed-temperature strategies, including the best average temperature and random selection, across various model-dataset combinations.

figures \ref{fig:temperature_distribution_AUROC}, \ref{fig:temperature_distribution_PR-AUC}, and \ref{fig:temperature_distribution_AURAC} present the AUROC, PR-AUC, and AURAC score distributions for different UQ methods across a range of fixed temperature values, complementing figure \ref{fig:semantic_entropy_AUROC} in the main text. These distributions highlight the significant impact of temperature selection on performance and underscore the limitations of static temperature choices.

figures \ref{fig:oracle_temperature_vs_mct_prauc} and \ref{fig:oracle_temperature_vs_mct_aurac} compare the performance of MCT with oracle-fixed temperature values using PR-AUC and AURAC metrics, complementing the results shown in figure \ref{fig:oracle_temperature_vs_mct_auroc}.

Tables \ref{tab:combined_results_prauc} and \ref{tab:combined_results_aurac} provide detailed performance comparisons for each UQ method across multiple models and datasets using the PR-AUC and AURAC metrics, complementing the results shown in Table \ref{tab:combined_results_auroc}.

\begin{table*}[ht!]
\centering
\renewcommand{\arraystretch}{1.0}
\setlength{\tabcolsep}{5pt}
\resizebox{0.8\textwidth}{!}{%
\begin{tabular}{|c|l|c|c|c|c|c|c|c|}
\hline
\multicolumn{9}{|c|}{\textbf{Discrete Semantic Entropy}} \\
\hline
\textbf{Model} & \textbf{Dataset} & \textbf{Oracle} & \textbf{MCT} & \textbf{Best Avg.} & \textbf{Random} & \textbf{MCT $\Delta$ (\%)} & \textbf{Best Avg. $\Delta$ (\%)}  & \textbf{Random $\Delta$ (\%)} \\
\hline
\multirow{4}{*}{\centering Falcon-7b-Instruct} & triviaQA & 0.7970 & 0.7899 & 0.7597 & 0.7738 & \textbf{0.89} & 4.68 & 2.92 \\
& \g bioasq & \g 0.8594 & \g 0.8325 & \g 0.8379 & \g 0.8333 & \g 3.13 & \g \textbf{2.51} & \g 3.04 \\
& svamp & 0.8145 & 0.8219 & 0.7937 & 0.8015 & \textbf{-0.90} & 2.56 & 1.60 \\
& \g nq & \g 0.8962 & \g 0.8987 & \g 0.8925 & \g 0.8866 & \g \textbf{-0.28} & \g 0.41 & \g 1.08 \\
\hline
\multirow{4}{*}{\centering Mistral-7b-Instruct} & triviaQA & 0.7057 & 0.6461 & 0.5665 & 0.6111 & \textbf{8.46} & 19.73 & 13.42 \\
& \g bioasq & \g 0.7493 & \g 0.7213 & \g 0.7129 & \g 0.7206 & \g \textbf{3.74} & \g 4.85 & \g 3.84 \\
& svamp & 0.6956 & 0.6974 & 0.6648 & 0.6496 & \textbf{-0.25} & 4.44 & 6.61 \\
& \g nq & \g 0.7893 & \g 0.7545 & \g 0.7547 & \g 0.7613 & \g 4.41 & \g 4.38 & \g \textbf{3.55} \\
\hline
\multirow{4}{*}{\centering Falcon-40b} & nq & 0.8103 & 0.7899 & 0.7637 & 0.7839 & \textbf{2.52} & 5.76 & 3.26 \\
& \g triviaQA & \g 0.6431 & \g 0.5978 & \g 0.5630 & \g 0.5745 & \g \textbf{7.05} & \g 12.47 & \g 10.68 \\
& svamp & 0.6812 & 0.6884 & 0.5426 & 0.5978 & \textbf{-1.06} & 20.35 & 12.25 \\
& \g bioasq & \g 0.7661 & \g 0.6980 & \g 0.6412 & \g 0.6836 & \g \textbf{8.89} & \g 16.31 & \g 10.77 \\
\hline
\multirow{4}{*}{\centering Llama-8b-Instruct} & triviaQA & 0.6817 & 0.6386 & 0.6427 & 0.6523 & 6.31 & 5.72 & \textbf{4.30} \\
& \g nq & \g 0.8100 & \g 0.8240 & \g 0.7921 & \g 0.7924 & \g \textbf{-1.73} & \g 2.21 & \g 2.17 \\
& bioasq & 0.6945 & 0.6668 & 0.6390 & 0.6568 & \textbf{3.98} & 7.98 & 5.42 \\
& \g svamp & \g 0.5701 & \g 0.6060 & \g 0.5618 & \g 0.5399 & \g \textbf{-6.28} & \g 1.46 & \g 5.30 \\
\hline
\multicolumn{9}{|c|}{\textbf{Naive Entropy}} \\
\hline
\multirow{4}{*}{\centering Falcon-7b-Instruct} & triviaQA & 0.7169 & 0.6753 & 0.6770 & 0.6758 & 5.80 & \textbf{5.56} & 5.73 \\
& \g bioasq & \g 0.7688 & \g 0.7540 & \g 0.7599 & \g 0.7440 & \g 1.92 & \g \textbf{1.16} & \g 3.22 \\
& svamp & 0.8058 & 0.7984 & 0.7746 & 0.7796 & \textbf{0.92} & 3.88 & 3.26 \\
& \g nq & \g 0.8826 & \g 0.8773 & \g 0.8783 & \g 0.8695 & \g 0.60 & \g \textbf{0.49} & \g 1.49 \\
\hline
\multirow{4}{*}{\centering Mistral-7b-Instruct} & triviaQA & 0.4536 & 0.3554 & 0.2993 & 0.3488 & \textbf{21.66} & 34.02 & 23.11 \\
& \g bioasq & \g 0.6634 & \g 0.6096 & \g 0.6474 & \g 0.6134 & \g 8.12 & \g \textbf{2.42} & \g 7.54 \\
& svamp & 0.5695 & 0.5928 & 0.5133 & 0.5161 & \textbf{-4.08} & 9.87 & 9.39 \\
& \g nq & \g 0.6832 & \g 0.6637 & \g 0.6627 & \g 0.6582 & \g \textbf{2.86} & \g 3.01 & \g 3.67 \\
\hline
\multirow{4}{*}{\centering Falcon-40b} & nq & 0.7034 & 0.6598 & 0.6864 & 0.6822 & 6.20 & \textbf{2.41} & 3.01 \\
& \g triviaQA & \g 0.4107 & \g 0.3730 & \g 0.3644 & \g 0.3746 & \g 9.17 & \g 11.28 & \g \textbf{8.80} \\
& svamp & 0.5710 & 0.5341 & 0.4634 & 0.5191 & \textbf{6.46} & 18.84 & 9.10 \\
& \g bioasq & \g 0.4988 & \g 0.4804 & \g 0.4580 & \g 0.4734 & \g \textbf{3.70} & \g 8.19 & \g 5.09 \\
\hline
\multicolumn{9}{|c|}{\textbf{Semantic Entropy}} \\
\hline
\multirow{4}{*}{\centering Falcon-7b-Instruct} & triviaQA & 0.7789 & 0.7767 & 0.7617 & 0.7473 & \textbf{0.28} & 2.21 & 4.06 \\
& \g bioasq & \g 0.8552 & \g 0.8247 & \g 0.8331 & \g 0.8193 & \g 3.57 & \g \textbf{2.58} & \g 4.19 \\
& svamp & 0.8029 & 0.8200 & 0.7845 & 0.7933 & \textbf{-2.12} & 2.29 & 1.21 \\
& \g nq & \g 0.8896 & \g 0.8908 & \g 0.8811 & \g 0.8779 & \g \textbf{-0.14} & \g 0.95 & \g 1.31 \\
\hline
\multirow{4}{*}{\centering Mistral-7b-Instruct} & triviaQA & 0.6788 & 0.5939 & 0.4749 & 0.5537 & \textbf{12.51} & 30.04 & 18.43 \\
& \g bioasq & \g 0.7218 & \g 0.6956 & \g 0.6755 & \g 0.6818 & \g \textbf{3.64} & \g 6.42 & \g 5.54 \\
& svamp & 0.6772 & 0.6591 & 0.6324 & 0.6172 & \textbf{2.69} & 6.62 & 8.87 \\
& \g nq & \g 0.7778 & \g 0.7395 & \g 0.7266 & \g 0.7413 & \g 4.92 & \g 6.58 & \g \textbf{4.68} \\
\hline
\multirow{4}{*}{\centering Falcon-40b} & nq & 0.8025 & 0.7909 & 0.7655 & 0.7809 & \textbf{1.44} & 4.62 & 2.69 \\
& \g triviaQA & \g 0.5964 & \g 0.5626 & \g 0.5634 & \g 0.5551 & \g 5.66 & \g \textbf{5.52} & \g 6.91 \\
& svamp & 0.6915 & 0.7190 & 0.5544 & 0.5896 & \textbf{-3.97} & 19.83 & 14.74 \\
& \g bioasq & \g 0.7572 & \g 0.7114 & \g 0.6187 & \g 0.6890 & \g \textbf{6.05} & \g 18.29 & \g 9.01 \\
\hline
\multicolumn{9}{|c|}{\textbf{Number of Semantic Sets}} \\
\hline
\multirow{4}{*}{\centering Falcon-7b-Instruct} & triviaQA & 0.7931 & 0.7871 & 0.7565 & 0.7681 & \textbf{0.76} & 4.62 & 3.16 \\
& \g bioasq & \g 0.8592 & \g 0.8341 & \g 0.8394 & \g 0.8386 & \g 2.92 & \g \textbf{2.31} & \g 2.40 \\
& svamp & 0.8136 & 0.8193 & 0.7922 & 0.8024 & \textbf{-0.70} & 2.64 & 1.38 \\
& \g nq & \g 0.8961 & \g 0.8992 & \g 0.8925 & \g 0.8889 & \g \textbf{-0.35} & \g 0.41 & \g 0.81 \\
\hline
\multirow{4}{*}{\centering Mistral-7b-Instruct} & triviaQA & 0.7053 & 0.6518 & 0.5646 & 0.6314 & \textbf{7.58} & 19.96 & 10.48 \\
& \g bioasq & \g 0.7470 & \g 0.7240 & \g 0.7168 & \g 0.7168 & \g \textbf{3.07} & \g 4.04 & \g 4.04 \\
& svamp & 0.6942 & 0.6984 & 0.6655 & 0.6553 & \textbf{-0.60} & 4.14 & 5.61 \\
& \g nq & \g 0.7904 & \g 0.7578 & \g 0.7594 & \g 0.7621 & \g 4.13 & \g 3.92 & \g \textbf{3.57} \\
\hline
\multirow{4}{*}{\centering Falcon-40b} & nq & 0.8105 & 0.7903 & 0.7630 & 0.7808 & \textbf{2.50} & 5.86 & 3.67 \\
& \g triviaQA & \g 0.6478 & \g 0.5994 & \g 0.5636 & \g 0.5773 & \g \textbf{7.47} & \g 13.01 & \g 10.88 \\
& svamp & 0.6748 & 0.6791 & 0.5408 & 0.5864 & \textbf{-0.64} & 19.85 & 13.11 \\
& \g bioasq & \g 0.7662 & \g 0.7001 & \g 0.6444 & \g 0.6894 & \g \textbf{8.62} & \g 15.89 & \g 10.01 \\
\hline
\multirow{4}{*}{\centering Llama-8b-Instruct} & triviaQA & 0.6884 & 0.6406 & 0.6490 & 0.6597 & 6.94 & 5.72 & \textbf{4.17} \\
& \g nq & \g 0.8100 & \g 0.8217 & \g 0.7940 & \g 0.7943 & \g \textbf{-1.45} & \g 1.98 & \g 1.95 \\
& bioasq & 0.6930 & 0.6726 & 0.6431 & 0.6609 & \textbf{2.94} & 7.20 & 4.64 \\
& \g svamp & \g 0.5671 & \g 0.6088 & \g 0.5659 & \g 0.5383 & \g \textbf{-7.36} & \g 0.21 & \g 5.07 \\
\hline
\multicolumn{9}{|c|}{\textbf{P(True)}} \\
\hline
\multirow{4}{*}{\centering Falcon-7b-Instruct} & triviaQA & 0.5504 & 0.5229 & 0.5380 & 0.5274 & 4.99 & \textbf{2.26} & 4.17 \\
& \g bioasq & \g 0.6931 & \g 0.5879 & \g 0.6425 & \g 0.6404 & \g 15.18 & \g \textbf{7.30} & \g 7.61 \\
& svamp & 0.6626 & 0.6399 & 0.6424 & 0.6480 & 3.43 & 3.05 & \textbf{2.19} \\
& \g nq & \g 0.8413 & \g 0.8280 & \g 0.8305 & \g 0.8255 & \g 1.58 & \g \textbf{1.29} & \g 1.89 \\
\hline
\multirow{4}{*}{\centering Mistral-7b-Instruct} & triviaQA & 0.6263 & 0.5140 & 0.5407 & 0.5748 & 17.93 & 13.67 & \textbf{8.22} \\
& \g bioasq & \g 0.7144 & \g 0.6528 & \g 0.6707 & \g 0.6708 & \g 8.63 & \g \textbf{6.11} & \g \textbf{6.11} \\
& svamp & 0.6216 & 0.5351 & 0.4813 & 0.5406 & 13.91 & 22.57 & \textbf{13.03} \\
& \g nq & \g 0.7955 & \g 0.7473 & \g 0.7602 & \g 0.7637 & \g 6.06 & \g 4.43 & \g \textbf{3.99} \\
\hline
\multirow{4}{*}{\centering Falcon-40b} & nq & 0.8417 & 0.7927 & 0.6954 & 0.7326 & \textbf{5.82} & 17.39 & 12.96 \\
& \g triviaQA & \g 0.5577 & \g 0.3654 & \g 0.4780 & \g 0.5111 & \g 34.49 & \g 14.29 & \g \textbf{8.35} \\
& svamp & 0.6386 & 0.6569 & 0.4888 & 0.5747 & \textbf{-2.88} & 23.45 & 10.00 \\
& \g bioasq & \g 0.7486 & \g 0.6228 & \g 0.6323 & \g 0.6832 & \g 16.80 & \g 15.54 & \g \textbf{8.73} \\
\hline
\end{tabular}%
}
\caption{Performance comparison of UQ methods using PR-AUC score. Bold values show best performance per scenario, with $\Delta$ indicating difference from oracle baseline (lower $\Delta$ is better). Note: MCT $\Delta$ may be negative when performance exceeds the oracle baseline.}
\label{tab:combined_results_prauc}
\end{table*}

\begin{table*}[ht!]
\centering
\renewcommand{\arraystretch}{1.0}
\setlength{\tabcolsep}{5pt}
\resizebox{0.8\textwidth}{!}{%
\begin{tabular}{|c|l|c|c|c|c|c|c|c|}
\hline
\multicolumn{9}{|c|}{\textbf{Discrete Semantic Entropy}} \\
\hline
\textbf{Model} & \textbf{Dataset} & \textbf{Oracle} & \textbf{MCT} & \textbf{Best Avg.} & \textbf{Random} & \textbf{MCT $\Delta$ (\%)} & \textbf{Best Avg. $\Delta$ (\%)}  & \textbf{Random $\Delta$ (\%)} \\
\hline
\multirow{4}{*}{\centering Falcon-7b-Instruct} & triviaQA & 0.6418 & 0.6295 & 0.6367 & 0.6046 & 1.91 & \textbf{0.80} & 5.79 \\
& \g bioasq & \g 0.5120 & \g 0.4878 & \g 0.5098 & \g 0.4833 & \g 4.72 & \g \textbf{0.42} & \g 5.60 \\
& svamp & 0.3592 & 0.3546 & 0.3341 & 0.3320 & \textbf{1.28} & 7.00 & 7.59 \\
& \g nq & \g 0.3304 & \g 0.3209 & \g 0.3265 & \g 0.3135 & \g 2.88 & \g \textbf{1.18} & \g 5.12 \\
\hline
\multirow{4}{*}{\centering Mistral-7b-Instruct} & triviaQA & 0.8448 & 0.8200 & 0.8137 & 0.8103 & \textbf{2.93} & 3.67 & 4.08 \\
& \g bioasq & \g 0.6608 & \g 0.6327 & \g 0.6567 & \g 0.6387 & \g 4.25 & \g \textbf{0.63} & \g 3.34 \\
& svamp & 0.7819 & 0.7576 & 0.7609 & 0.7427 & 3.11 & \textbf{2.70} & 5.01 \\
& \g nq & \g 0.5747 & \g 0.5557 & \g 0.5502 & \g 0.5488 & \g \textbf{3.30} & \g 4.26 & \g 4.51 \\
\hline
\multirow{4}{*}{\centering Falcon-40b} & nq & 0.5261 & 0.5245 & 0.5104 & 0.5127 & \textbf{0.32} & 2.99 & 2.56 \\
& \g triviaQA & \g 0.8703 & \g 0.8738 & \g 0.8619 & \g 0.8569 & \g \textbf{-0.40} & \g 0.97 & \g 1.53 \\
& svamp & 0.6853 & 0.6654 & 0.6258 & 0.6480 & \textbf{2.90} & 8.69 & 5.46 \\
& \g bioasq & \g 0.6494 & \g 0.6430 & \g 0.6115 & \g 0.6297 & \g \textbf{1.00} & \g 5.84 & \g 3.03 \\
\hline
\multirow{4}{*}{\centering Llama-8b-Instruct} & triviaQA & 0.8298 & 0.8149 & 0.8222 & 0.8183 & 1.79 & \textbf{0.91} & 1.38 \\
& \g nq & \g 0.5649 & \g 0.5565 & \g 0.5634 & \g 0.5558 & \g 1.49 & \g \textbf{0.27} & \g 1.60 \\
& bioasq & 0.7121 & 0.6934 & 0.6961 & 0.6971 & 2.62 & 2.25 & \textbf{2.11} \\
& \g svamp & \g 0.7042 & \g 0.7290 & \g 0.7042 & \g 0.6863 & \g \textbf{-3.52} & \g 0.00 & \g 2.55 \\
\hline
\multicolumn{9}{|c|}{\textbf{Naive Entropy}} \\
\hline
\multirow{4}{*}{\centering Falcon-7b-Instruct} & triviaQA & 0.6314 & 0.6029 & 0.6001 & 0.6056 & 4.51 & 4.96 & \textbf{4.08} \\
& \g bioasq & \g 0.4968 & \g 0.4867 & \g 0.4932 & \g 0.4858 & \g 2.03 & \g \textbf{0.72} & \g 2.22 \\
& svamp & 0.3623 & 0.3652 & 0.3423 & 0.3492 & \textbf{-0.82} & 5.50 & 3.59 \\
& \g nq & \g 0.3369 & \g 0.3287 & \g 0.3320 & \g 0.3218 & \g 2.43 & \g \textbf{1.43} & \g 4.47 \\
\hline
\multirow{4}{*}{\centering Mistral-7b-Instruct} & triviaQA & 0.8278 & 0.8085 & 0.8191 & 0.8022 & 2.33 & \textbf{1.05} & 3.09 \\
& \g bioasq & \g 0.6805 & \g 0.6631 & \g 0.6699 & \g 0.6601 & \g 2.56 & \g \textbf{1.56} & \g 3.00 \\
& svamp & 0.7608 & 0.7573 & 0.7550 & 0.7466 & \textbf{0.46} & 0.76 & 1.88 \\
& \g nq & \g 0.5618 & \g 0.5554 & \g 0.5537 & \g 0.5500 & \g \textbf{1.14} & \g 1.44 & \g 2.09 \\
\hline
\multirow{4}{*}{\centering Falcon-40b} & nq & 0.5166 & 0.5159 & 0.5049 & 0.5087 & \textbf{0.14} & 2.28 & 1.52 \\
& \g triviaQA & \g 0.8823 & \g 0.8730 & \g 0.8735 & \g 0.8724 & \g 1.05 & \g \textbf{1.00} & \g 1.12 \\
& svamp & 0.6708 & 0.6276 & 0.6270 & 0.6460 & 6.44 & 6.53 & \textbf{3.69} \\
& \g bioasq & \g 0.5938 & \g 0.5872 & \g 0.5653 & \g 0.5765 & \g \textbf{1.10} & \g 4.79 & \g 2.91 \\
\hline
\multicolumn{9}{|c|}{\textbf{Semantic Entropy}} \\
\hline
\multirow{4}{*}{\centering Falcon-7b-Instruct} & triviaQA & 0.6637 & 0.6464 & 0.6356 & 0.6074 & \textbf{2.62} & 4.24 & 8.48 \\
& \g bioasq & \g 0.5266 & \g 0.5036 & \g 0.5233 & \g 0.4942 & \g 4.35 & \g \textbf{0.63} & \g 6.15 \\
& svamp & 0.3680 & 0.3728 & 0.3462 & 0.3471 & \textbf{-1.29} & 5.93 & 5.69 \\
& \g nq & \g 0.3458 & \g 0.3308 & \g 0.3458 & \g 0.3242 & \g 4.33 & \g \textbf{0.00} & \g 6.25 \\
\hline
\multirow{4}{*}{\centering Mistral-7b-Instruct} & triviaQA & 0.8522 & 0.8215 & 0.8428 & 0.8148 & 3.60 & \textbf{1.10} & 4.38 \\
& \g bioasq & \g 0.6650 & \g 0.6390 & \g 0.6617 & \g 0.6440 & \g 3.90 & \g \textbf{0.49} & \g 3.16 \\
& svamp & 0.7990 & 0.7584 & 0.7700 & 0.7573 & 5.08 & \textbf{3.63} & 5.22 \\
& \g nq & \g 0.5829 & \g 0.5685 & \g 0.5825 & \g 0.5554 & \g 2.47 & \g \textbf{0.07} & \g 4.73 \\
\hline
\multirow{4}{*}{\centering Falcon-40b} & nq & 0.5466 & 0.5350 & 0.5321 & 0.5300 & \textbf{2.12} & 2.66 & 3.03 \\
& \g triviaQA & \g 0.8827 & \g 0.8730 & \g 0.8688 & \g 0.8587 & \g \textbf{1.10} & \g 1.57 & \g 2.72 \\
& svamp & 0.7050 & 0.7012 & 0.6491 & 0.6680 & \textbf{0.54} & 7.92 & 5.25 \\
& \g bioasq & \g 0.6622 & \g 0.6632 & \g 0.6284 & \g 0.6449 & \g \textbf{-0.15} & \g 5.11 & \g 2.62 \\
\hline
\multicolumn{9}{|c|}{\textbf{Number of Semantic Sets}} \\
\hline
\multirow{4}{*}{\centering Falcon-7b-Instruct} & triviaQA & 0.6320 & 0.6218 & 0.6084 & 0.5912 & \textbf{1.61} & 3.73 & 6.46 \\
& \g bioasq & \g 0.5042 & \g 0.4834 & \g 0.5042 & \g 0.4850 & \g 4.13 & \g \textbf{0.00} & \g 3.81 \\
& svamp & 0.3511 & 0.3475 & 0.3246 & 0.3293 & \textbf{1.03} & 7.55 & 6.20 \\
& \g nq & \g 0.3240 & \g 0.3140 & \g 0.3221 & \g 0.3100 & \g 3.08 & \g \textbf{0.57} & \g 4.30 \\
\hline
\multirow{4}{*}{\centering Mistral-7b-Instruct} & triviaQA & 0.8420 & 0.8187 & 0.8118 & 0.8158 & \textbf{2.77} & 3.59 & 3.12 \\
& \g bioasq & \g 0.6567 & \g 0.6312 & \g 0.6516 & \g 0.6302 & \g 3.89 & \g \textbf{0.78} & \g 4.04 \\
& svamp & 0.7783 & 0.7548 & 0.7581 & 0.7469 & 3.02 & \textbf{2.59} & 4.04 \\
& \g nq & \g 0.5687 & \g 0.5512 & \g 0.5478 & \g 0.5435 & \g \textbf{3.06} & \g 3.67 & \g 4.42 \\
\hline
\multirow{4}{*}{\centering Falcon-40b} & nq & 0.5200 & 0.5197 & 0.5025 & 0.5068 & \textbf{0.06} & 3.38 & 2.54 \\
& \g triviaQA & \g 0.8669 & \g 0.8715 & \g 0.8592 & \g 0.8550 & \g \textbf{-0.54} & \g 0.89 & \g 1.37 \\
& svamp & 0.6759 & 0.6509 & 0.6134 & 0.6357 & \textbf{3.70} & 9.24 & 5.94 \\
& \g bioasq & \g 0.6456 & \g 0.6349 & \g 0.6035 & \g 0.6246 & \g \textbf{1.66} & \g 6.52 & \g 3.26 \\
\hline
\multirow{4}{*}{\centering Llama-8b-Instruct} & triviaQA & 0.8265 & 0.8120 & 0.8223 & 0.8164 & 1.76 & \textbf{0.51} & 1.23 \\
& \g nq & \g 0.5582 & \g 0.5459 & \g 0.5575 & \g 0.5498 & \g 2.22 & \g \textbf{0.13} & \g 1.51 \\
& bioasq & 0.7052 & 0.6887 & 0.6957 & 0.6920 & 2.35 & \textbf{1.36} & 1.87 \\
& \g svamp & \g 0.6965 & \g 0.7233 & \g 0.6965 & \g 0.6820 & \g \textbf{-3.85} & \g 0.00 & \g 2.08 \\
\hline
\multicolumn{9}{|c|}{\textbf{P(True)}} \\
\hline
\multirow{4}{*}{\centering Falcon-7b-Instruct} & triviaQA & 0.4866 & 0.4485 & 0.4547 & 0.4543 & 7.82 & \textbf{6.54} & 6.63 \\
& \g bioasq & \g 0.4436 & \g 0.3315 & \g 0.4009 & \g 0.3975 & \g 25.26 & \g \textbf{9.62} & \g 10.38 \\
& svamp & 0.2340 & 0.2121 & 0.2340 & 0.2178 & 9.37 & \textbf{0.00} & 6.93 \\
& \g nq & \g 0.2953 & \g 0.2673 & \g 0.2747 & \g 0.2773 & \g 9.48 & \g 7.00 & \g \textbf{6.10} \\
\hline
\multirow{4}{*}{\centering Mistral-7b-Instruct} & triviaQA & 0.8542 & 0.8272 & 0.8282 & 0.8376 & 3.17 & 3.04 & \textbf{1.94} \\
& \g bioasq & \g 0.7296 & \g 0.7012 & \g 0.7067 & \g 0.7059 & \g 3.90 & \g \textbf{3.13} & \g 3.25 \\
& svamp & 0.7532 & 0.7207 & 0.7274 & 0.7359 & 4.31 & 3.42 & \textbf{2.29} \\
& \g nq & \g 0.6109 & \g 0.5832 & \g 0.5925 & \g 0.5936 & \g 4.54 & \g 3.00 & \g \textbf{2.84} \\
\hline
\multirow{4}{*}{\centering Falcon-40b} & nq & 0.5519 & 0.5463 & 0.5100 & 0.5231 & \textbf{1.00} & 7.59 & 5.22 \\
& \g triviaQA & \g 0.8844 & \g 0.8652 & \g 0.8759 & \g 0.8718 & \g 2.17 & \g \textbf{0.95} & \g 1.42 \\
& svamp & 0.6819 & 0.6272 & 0.6091 & 0.6496 & 8.03 & 10.69 & \textbf{4.74} \\
& \g bioasq & \g 0.7084 & \g 0.6782 & \g 0.6771 & \g 0.6901 & \g 4.26 & \g 4.41 & \g \textbf{2.58} \\
\hline
\end{tabular}%
}
\caption{Performance comparison of UQ methods using AURAC score. Bold values show best performance per scenario, with $\Delta$ indicating difference from oracle baseline (lower $\Delta$ is better). Note: MCT $\Delta$ may be negative when performance exceeds the oracle baseline}
\label{tab:combined_results_aurac}
\end{table*}

\end{document}